\documentclass{article}
\usepackage{ijcai24}

\usepackage{times}
\usepackage{soul}
\usepackage{url}
\usepackage[hidelinks]{hyperref}
\usepackage[utf8]{inputenc}
\usepackage[small]{caption}
\usepackage{graphicx}
\usepackage{amsmath}
\usepackage{amsthm}
\usepackage{booktabs}
\usepackage{algorithm}
\usepackage{algorithmic}
\usepackage[switch]{lineno}

\usepackage{tabularx}
\usepackage{graphbox}
\usepackage{multirow}

\title{Beyond Direct Diagnosis: LLM-based Multi-Specialist Agent Consultation for Automatic Diagnosis}

\author {
    Haochun Wang \and Sendong Zhao\thanks{Corresponding author} \and Zewen Qiang \and Nuwa Xi \and Bing Qin \and Ting Liu
\affiliations
   Research Center for Social Computing and Information Retrieval, \\Harbin Institute of Technology, China
\emails    \{hcwang, sdzhao\}@ir.hit.edu.cn
}

\begin{document}

\maketitle

\begin{abstract}
Automatic diagnosis is a significant application of AI in healthcare, where diagnoses are generated based on the symptom description of patients. Previous works have approached this task directly by modeling the relationship between the normalized symptoms and all possible diseases. However, in the clinical diagnostic process, patients are initially consulted by a general practitioner and, if necessary, referred to specialists in specific domains for a more comprehensive evaluation. The final diagnosis often emerges from a collaborative consultation among medical specialist groups. Recently, large language models have shown impressive capabilities in natural language understanding. In this study, we adopt tuning-free LLM-based agents as medical practitioners and propose the \textbf{A}gent-derived \textbf{M}ulti-\textbf{S}pecialist \textbf{C}onsultation (AMSC) framework to model the diagnosis process in the real world by adaptively fusing probability distributions of agents over potential diseases. Experimental results demonstrate the superiority of our approach compared with baselines. Notably, our approach requires significantly less parameter updating and training time, enhancing efficiency and practical utility. Furthermore, we delve into a novel perspective on the role of implicit symptoms within the context of automatic diagnosis. 
\end{abstract}

\section{Introduction}
The utilization of artificial intelligence in healthcare, specifically in the realm of automatic diagnosis, has represented a pivotal advancement. Healthcare agents towards automatic diagnosis are designed to interact with patients, eliciting information about symptom presence and formulating diagnoses based on these patient-reported discomforts, termed as complaints. Such AI-driven automatic diagnostic systems are poised to assume a critical role in primary healthcare, offering the potential to mitigate the challenges posed by medical resource scarcity, particularly in economically disadvantaged regions \cite{wang2023coad,world2016health}.

\begin{figure}[ht] 
\centering 
\includegraphics[width=0.98\columnwidth]{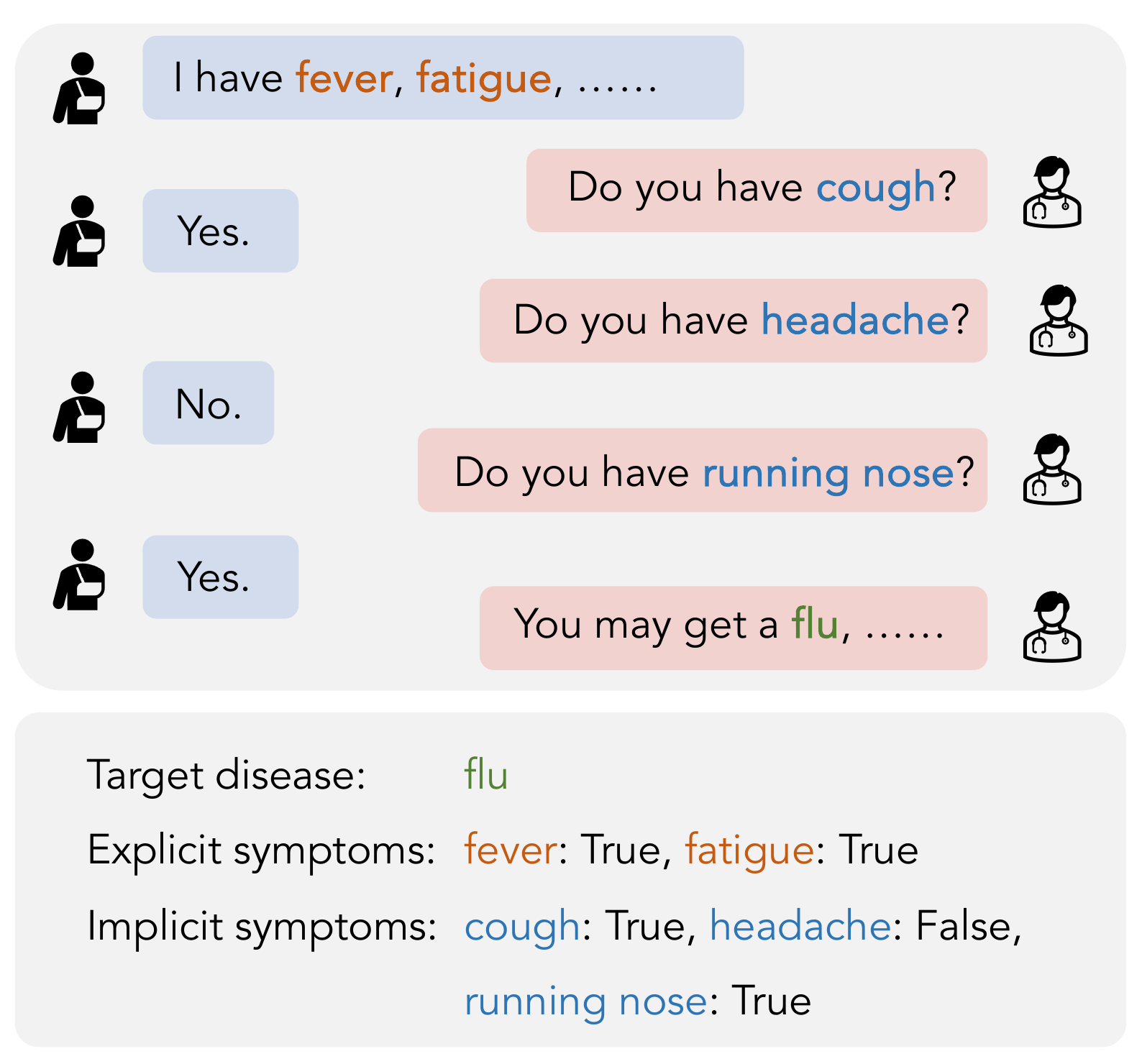} 
\caption{An example of data for automatic diagnosis from online consulting platforms.} 
\label{intro_task} 
\end{figure}

The task of automatic diagnosis is derived from the process in real-world online consultation platforms, which is depicted in Figure \ref{intro_task}. It involves the annotation of dialogue data between the patients and practitioners including three distinct components: (1) Explicit Symptoms - those symptoms presented in the initial patient complaint; (2) Implicit Symptoms - additional symptoms ascertained through subsequent practitioner inquiries; (3) Target Disease - the diagnosis predicted by the practitioner, based on a comprehensive assessment of symptoms. Previous research predominantly addresses the automatic diagnosis task by initially predicting implicit symptoms using explicit symptoms. Subsequently, a diagnosis is formulated leveraging all symptoms, employing methodologies such as reinforcement learning \cite{tang2016inquire,wei2018task,hou2021imperfect} or sequence generation \cite{chen2022diaformer,wang2023coad}. Existing approaches tend to model the dynamics of symptom-disease relationships but ignore the diagnosis process in real-life clinical settings.

\begin{figure*}[ht] 
\centering 
\includegraphics[width=\textwidth]{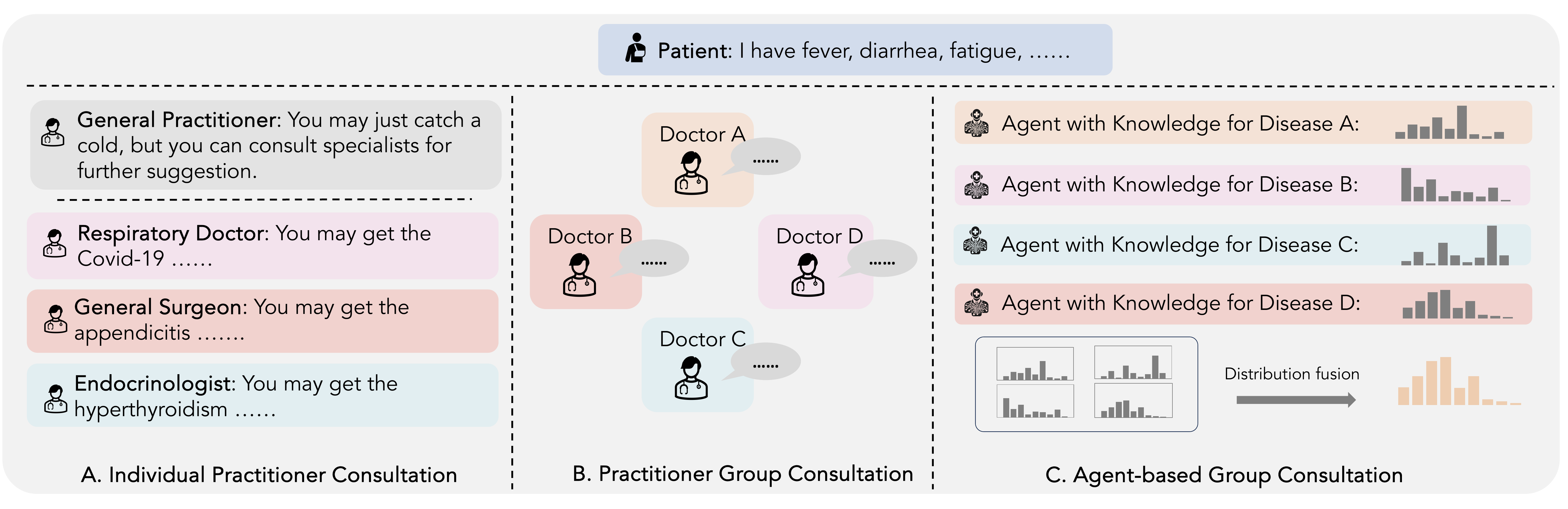} 
\caption{Scenarios of three medical diagnostic processes. A. Individual Practitioner Consultation: a singular general practitioner or specialist formulates the diagnosis; B. Practitioner Group Consultation: a group of professionals collaboratively arriving at a diagnostic conclusion; C. Agent-based Group Consultation: the diagnosis is derived from the decision fusion from multiple agent-based specialists.} 
\label{intro_case} 
\end{figure*}

In clinical practice, when a patient presents with symptoms necessitating medical intervention, the customary procedure commences with an initial evaluation by a general practitioner. This primary assessment aims to establish a preliminary diagnosis. Subsequently, the general practitioner typically refers the patient to a specialist with expertise in the relevant medical domain for advanced consultation and management if necessary, as depicted in Scenario A of Figure \ref{intro_case}. Furthermore, for instances involving multifaceted medical conditions, it is standard to convene a multidisciplinary team of practitioners, each contributing diverse expertise and experience. This collaborative approach facilitates a holistic assessment and culminates in a consensus diagnosis, as illustrated in Scenario B of Figure \ref{intro_case}.

In recent advancements, Large Language Models (LLMs) have demonstrated exceptional capabilities in both comprehending and generating natural language across diverse tasks \cite{brown2020language,touvron2023llama,chowdhery2023palm,openai2023gpt4}. Moreover, their application within the medical field has garnered significant research interest \cite{wang2023huatuo,bao2023disc}. Building upon this foundation, this study introduces the \textbf{A}gent-derived \textbf{M}ulti-\textbf{S}pecialist \textbf{C}onsultation (AMSC) model, conceptualized from real-world clinical scenarios. In this model, we envision open-source LLMs as primary care physicians, while specialized LLM-based agents, each equipped with distinct medical knowledge, function akin to medical specialists. These agents, each possessing expertise in different medical domains, generate predictive distributions for possible diseases or medical categories without parameter updating of LLMs. Subsequently, we propose an adaptive probability distribution fusion method that integrates the agent predictions, offering a refined decision-making process for disease diagnosis. Additionally, we scrutinize the influence of implicit symptoms on automated diagnosis, investigating the impact of various symptoms on the diagnostic outcomes.

In summary, our contributions are outlined as follows: 
\begin{itemize}
    \item We introduce an agent-based multi-specialist consultation model employing open-source LLMs following real-world clinical scenarios, coupled with an adaptive predictive distribution fusion method for disease diagnosis.
    \item Experimental results demonstrate that our approach surpasses existing baselines in performance. Notably, as the model only requires parameter training on the distribution fusion, it is also more efficient in terms of training costs.
    \item We also conduct a comprehensive analysis of symptomatology in automated diagnosis, exploring the differential impacts of various symptoms on diagnostic results.
\end{itemize}

\section{Related Work}
\subsection{Automatic Diagnosis}
Automatic diagnosis using reinforcement learning (RL) has evolved significantly. Initial work by \cite{tang2016inquire} used RL for neural symptom checking, followed by hierarchical RL approaches for joint diagnostics and symptom inquiry \cite{kao2018context,liao2020task}. \cite{wei2018task} introduced a Deep Q-Network for symptom collection, while \cite{xu2019end} emphasized integrating prior medical knowledge into policy learning. Further advancements include decoupling symptoms and diseases using dense representations \cite{chen2023dxformer}, policy gradient frameworks \cite{xia2020generative}, multi-level reward models \cite{hou2021imperfect}, and dialog data customization \cite{teixeira2021interplay}. Beyond RL, \cite{chen2022diaformer} and \cite{hou2023mtdiag} explored sequence generation and multi-task frameworks for diagnosis, respectively. \cite{wang2023coad} proposed a joint symptom-disease generation model. However, existing research lacks focus on cross-specialist knowledge integration in automatic diagnosis. Our study addresses this gap by exploring the fusion of diagnoses with agent specialists.

\begin{figure*}[ht] 
\centering 
\includegraphics[width=\textwidth]{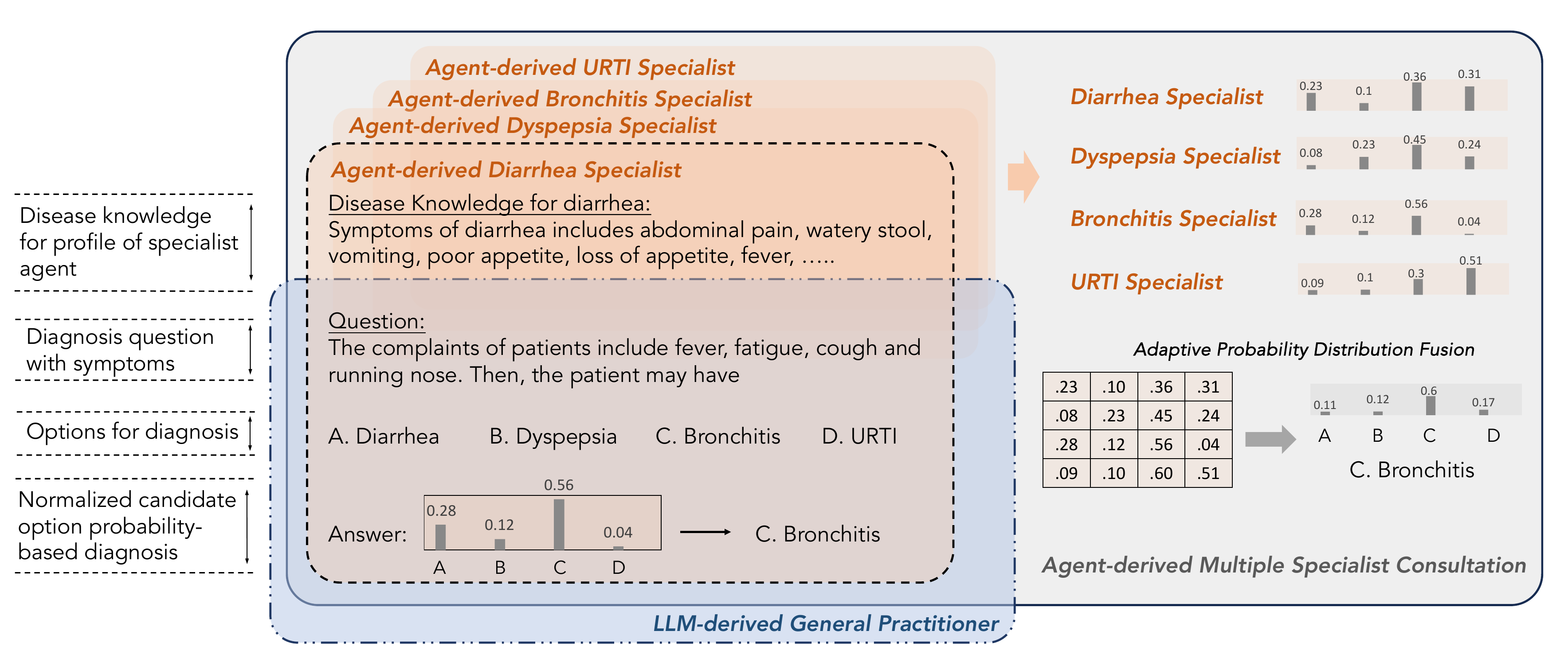} 
\caption{Illustration of LLM-derived General Practitioner, Agent-derived Specialist and Agent-derived Multiple Specialist Consultation.} 
\label{methodology} 
\end{figure*}

\subsection{Medical Large Language Models}
Recent study has underscored the efficacy of Large Language Models (LLMs) within the medical field~\cite{thirunavukarasu2023large,bao2023disc}, manifesting in diverse applications, including diagnostic processes \cite{singhal2023large}, medical reasoning \cite{lievin2022can}, and the summarization of medical evidence \cite{tang2023evaluating}. In the context of knowledge-augmented methodologies, \cite{wang2023knowledge} demonstrated the generation of informed responses via queries to knowledge bases. Furthermore, \cite{kang2023knowledge} fine-tuned smaller LLMs with rationales derived from retrieved knowledge. Another notable advancement is \cite{jin2023genegpt} GeneGPT, which instructs LLMs to integrate Web APIs, thereby enhancing performance. Despite these advancements, the medical application faces inherent privacy challenges, particularly concerning the uploading of personal health information to online LLMs. To address this, we introduce the AMSC framework, tailored for open-source LLMs and deployed locally and privately without privacy concern.

\section{Agent-derived Multi-specialist Consultation Framework}
In this section, we delve into an Agent-derived Multi-Specialist Consultation (AMSC) framework, including refining the task of automatic diagnosis, implementing an agent-derived specialist, and finally describing the fusion of multi-specialist decisions derived from these agents in medical diagnostics mimicking the procedure in the real-life diagnosis by multiple specialists.
\subsection{Refinement of Automatic Diagnosis}
The task of automatic diagnosis involves processing patient-specific data, denoted as $p_i$. This data encompasses the target disease $d_i$, an array of explicit symptoms $s_{i}^{exp}=\{exp_1,exp_2,......\}$, and a collection of implicit symptoms $s_{i}^{imp}=\{imp_1,imp_2,......\}$. Explicit symptoms are direct manifestations reported by patients, while implicit symptoms are deduced from interactive dialogues between patients and practitioners. Both categories of symptoms undergo standardization to medical terminologies from their original natural language descriptions.

Recent advancements in LLMs have demonstrated remarkable efficacy in natural language processing within medical contexts \cite{bao2023disc,wang2023huatuo}. In our approach, we leverage these LLMs as virtual medical practitioners. Given the finite nature of disease possibilities in automatic diagnosis tasks, we restructure the task into a Multiple-Choice Question-Answering (MCQA) framework, as illustrated in Figure \ref{methodology}, which involves constructing symptom-based question templates and selecting the results from answer options, formulated as  

\begin{equation*}
p_{disease_{i}} = \frac{\texttt{LLM}\left(q,\ opts;i\right)}{\sum_{j} \texttt{LLM}\left(q,\ opts;j\right)}
\end{equation*}
where $q$ represents the symptom-inclusive query question for the LLM, $opts$ denotes the array of possible disease diagnoses, \texttt{LLM} refers to the process that large language model generates the probability of each token within its vocabulary and $i,j$ are indices corresponding to the specific token options for disease diagnosis.

For tasks involving a broader array of choices, we employ a hierarchical MCQA structure based on ICD-10-CM codes \cite{world2004international}. Meanwhile, recent research demonstrates that the effectiveness of MCQAs in leveraging the capability of LLMs is subject to the Multiple-Choice Symbol Binding (MCSB) proficiency of the model (the capability of LLMs to associate the answer options with the alphabetic symbols), which significantly varies across different models \cite{robinson2023leveraging,zheng2023large}. Therefore, we selectively utilize models with superior multiple-choice symbol binding capabilities and empirically evaluate their Proportion of Plurality Agreement (PPA) value \cite{robinson2023leveraging} (further details are provided in the experimental section).

\subsection{Agent-derived Specialist for Diagnosis}
While LLM-based practitioners possess a foundational understanding of medical knowledge akin to general practitioners, they lack the depth of expertise characteristic of specialists in particular diseases. To address this, we enhance the LLMs with specialized disease knowledge profiles from the National Institutes of Health\footnote{https://www.nhlbi.nih.gov/health}, thereby transforming them into distinct specialist agents as depicted in Figure \ref{methodology}. Each agent-specialist is endowed with expertise in a singular disease domain. For a question for diagnosis, an agent-specialist outputs a distribution on all the possible answer options, and the corresponding disease for the option with the highest probability is the diagnostic result, denoted as

\begin{equation*}
k_n = d_n , (sym_{n_1},sym_{n_2}, ...)
\end{equation*}

\begin{equation*}
\texttt{Agent}_n = \texttt{Agent-specialist}(\texttt{LLM}, k_n)
\end{equation*}

\begin{equation*}
p_{disease_{i}}^n = \frac{\texttt{Agent}_n \left(q,\ opts;i\right)}{\sum_{j} \texttt{Agent}_n\left(q,\ opts;j\right)}
\end{equation*}
where a disease knowledge $k_n$ includes a disease $d_n$ and a series of symptoms $(sym_{n_1},sym_{n_2}, ...)$ of the disease $d_n$. $\texttt{Agent}_n$ represents an agent-derived specialist with a LLM and specific disease knowledge $k_n$.

\begin{figure}[ht] 
\includegraphics[width=0.98\columnwidth]{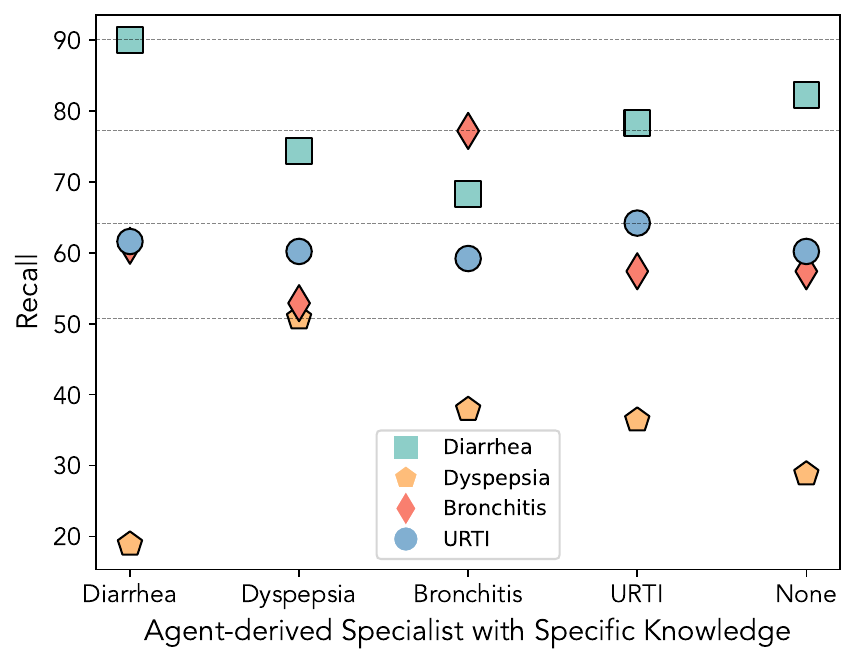} 
\caption{Recall for the four agent-derived specialists on the diseases of the MuZhi-4 dataset with corresponding knowledge for a specific disease and for an LLM-derived general practitioner without knowledge (``None'' in the figure).} 
\label{recall_analysis} 
\end{figure}

To assess whether an agent specialist exhibits enhanced knowledge in its respective disease area as a human specialist, we conduct experiments on the most widely used automatic diagnosis dataset MuZhi-4 \cite{wei2018task}, which includes four target diseases. Figure \ref{recall_analysis} presents the recall value achieved by these agent specialists for each disease, using distinct symbols. The results indicate that an agent with specialized knowledge in a particular disease significantly outperforms those with knowledge of other diseases and the LLM-based general practitioner without such specialization, affirming the capability of agent-derived specialists to effectively assimilate and apply specialized knowledge.

\subsection{Multi-Specialist Decision Fusion}
Like in the real world where multiple specialists may yield different diagnoses for the same patient, in the realm of the Agent-derived Multi-Specialist Consultation (AMSC) framework, each agent conceptualized as a specialist, yields a probabilistic distribution over a range of potential diagnoses. This process results in the formation of a diagnosis distribution matrix when multiple agents are involved. Analogous to the consensus-building process among human medical specialists through extensive deliberation, we introduce an innovative approach for reaching a conclusive diagnostic decision from this fusion of agent-specialist inputs. Our methodology employs a self-attention-based Adaptive Probability Distribution Fusion (APDF) technique. 

We consider a diagnosis distribution matrix denoted as $M=\{m_1,m_2, ...\}$, where $m_n$ represents the diagnostic probability distribution contributed by the $n^{\text{th}}$ agent-specialist. The following equations delineate our method:

\begin{equation*}
    Q,K,V = \texttt{Linear}_{q,k,v}(M) 
\end{equation*}

\begin{equation*}
   p_{disease_{i}} = \texttt{Linear}(softmax(\frac{QK^T}{\sqrt{d_k}})V;i)
\end{equation*}

Here, $Q,K,V$ represent the linear projections of the distribution matrix $M$ corresponding to the query, key, and value components respectively, utilizing the transformation $\texttt{Linear}_{q,k,v}$. The final diagnostic outcome is derived from an adaptively weighted fusion of the diagnostic distributions sourced from multiple agent specialists, capitalizing on the self-attention mechanism's capacity to discern and emphasize relevant features within the distribution matrix.
\section{Experiment}
\subsection{Datasets}
We evaluate our approach on three datasets derived from real-life contexts: 
\begin{itemize}
\item \textbf{MuZhi-4} \cite{wei2018task}, a meticulously annotated dataset collected from an online platform Baidu MuZhi Doctor \footnote{https://muzhi.baidu.com/} which focuses on four diseases in the pediatric domain; 
\item \textbf{MuZhi-10} is an augmented iteration of the MuZhi-4 dataset, extending to ten diseases, spanning respiratory, digestive and endocrine systems; 
\item \textbf{Dxy} \cite{xu2019end} is a medical dialogue dataset curated from Dingxiang Doctor platform. \footnote{https://dxy.com}
\end{itemize}

A thorough statistical analysis of the datasets is presented in Table \ref{statistics}. Additionally, prior research has engaged in experimental analysis utilizing synthetic datasets \cite{liao2020task,chen2022diaformer}  fabricated by symptom-disease database Symcat \footnote{https://www.symcat.com}. Notably, as depicted in Table \ref{statistics}, the datasets derived from real-life contexts exhibit an average of 2.36 explicit symptoms per patient complaint. In stark contrast, the synthetic dataset only contains 1 explicit symptom per case. This indicates that during medical consultations, patients tend to articulate their ailments with greater specificity in real-life scenarios, as opposed to the limited expressiveness observed in synthetic datasets. Such a discrepancy underscores a significant divergence in the distribution of data between synthetic and real-life datasets. Consequently, our study focuses exclusively on datasets derived from real-life scenarios, thereby ensuring a more realistic and applicable analysis.

\begin{table}[htbp]
\renewcommand{\arraystretch}{1.15}
\begin{tabular}{lccc|c}
\hline
 & \textbf{MZ-4} & \textbf{MZ-10} & \textbf{Dxy} & \textbf{Synthetic} \\ \hline
\# Disease     & 4           & 10             & 5            &  90               \\
\# Symptom     & 66          & 331             & 41           &  266                \\ \hline
\# Training    & 568         & 3704          & 423          &  24000             \\
\# Tr\_avg\_exp & 2.39       & 1.72           & 3.09         &  1.00            \\
\# Tr\_avg\_imp & 3.28       & 5.29           & 1.65         &  2.61              \\ \hline
\# Test        & 142         & 412           & 104          &  6000          \\
\# T\_avg\_exp  & 2.20    &  1.77           &  3.00         &  1.00              \\
\# T\_avg\_imp  & 3.20    &  5.18           & 1.76         &  2.59             \\
\hline
\end{tabular}
\caption{Statistics of the datasets, including three real-life datasets and one synthetic dataset. \#Tr\_avg\_exp: the average number of explicit symptoms in the training set; \#Tr\_avg\_imp: the average number of implicit symptoms in the training set; \#T\_avg\_exp: the average number of explicit symptoms in the test set; \#T\_avg\_imp: the average number of implicit symptoms in the test set.}
\label{statistics}
\end{table}

\begin{table*}[htbp]
\centering
\renewcommand{\arraystretch}{1.15}
\resizebox{\textwidth}{!}{
\begin{tabular}{lcccccccccccc}
\hline
\multirow{2}{*}{\textbf{Model}} & \multicolumn{4}{c}{MuZhi-4} & \multicolumn{4}{c}{Dxy}  & \multicolumn{4}{c}{MuZhi-10} \\
\cmidrule(l){2-5} \cmidrule(l){6-9} \cmidrule(l){10-13} 
& Acc & Ttime & \#Params & Turn & Acc & Ttime & \#Params & Turn& Acc & Ttime & \#Params & Turn \\ \hline
DQN       & 0.690 & 82m  & -   & 2.0  & 0.720 & 141m &  -  & 1.3  & 0.408 & -    & - & 9.7   \\
HRL       & 0.694 & 162m & -   & 3.5  & 0.695 & 35m  &  -  & 2.4  & 0.556 & -    & - & 7.0\\
PPO       & 0.732 &   -  & -   & 6.3  & 0.746 &  -   &  -  & 3.3  &  -    & -    & - & - \\
Diaformer & 0.742 & 2m   & 62M & 15.3 & 0.829 &  2m  & 62M & 13.1 &  -    & -    & - & - \\
CoAD      & 0.75  & 32m  & 43M & 13.4 & 0.85  &  23m & 43M & 10.5 & 0.628 & 156m & 43M  & 20.0 \\
DxFormer  & 0.743 & 39m  & 9.8M & 8.7 & 0.817 &  14m & 9.8M &6.3&0.633& 55m & 9.8M & 7.6 \\ \hline
AMSC \dag & \textbf{0.773}   &  0.15m   &  832  &  0   &  \textbf{0.861} & 0.09m & 2000 &  0  &  \textbf{0.652}  &  9m  & 0.031M &  0  \\ 
\ \ -Single agent \dag\ddag & 0.676 & 0m & 0 & 0 & 0.776 & 0m & 0 & 0 &  0.506     & 0m & 0 & 0 \\ 
\ \ -LLM \dag               & 0.632 & 0m & 0 & 0 & 0.724 & 0m & 0 & 0 &  0.427     & 0m & 0 & 0 \\ \hline  
\end{tabular}
}
\caption{Experimental results on three real-life datasets. Acc is the accuracy of diagnosis; Ttime denotes the training time, m for minute; \#Params indicates the number of updated parameters, M for million; Turn represents the mean turn of symptom inquiry. The results of accuracy are reported by baselines. Single agent: the performance of agent-derived specialist without distribution fusion (\dag: with no implicit symptom. \ddag: best performance of multiple agents). LLM: the performance of original LLM.}
\label{main_result}
\end{table*}
\subsection{Baselines}
In this research, we present an analytical comparison of our experimental results with a spectrum of established baseline models in the domain of automatic medical diagnosis. These benchmarks include: 
\begin{itemize}
    \item DQN model \cite{wei2018task}, which employs a Deep Q-Network framework for symptom validation and diagnostic processes.
    \item HRL approach \cite{liao2020task}, which implements a DQN-based hierarchical policy to separately predict symptoms and diagnose diseases.
    \item PPO model \cite{schulman2017proximal}, which utilizes proximal policy optimization techniques for the development of a diagnostic model.
    \item DxFormer \cite{chen2023dxformer}, a novel model that decouples the processes of symptom inquiry and disease diagnosis through dense symptom representation.
    \item Diaformer \cite{chen2022diaformer}, which conceptualizes the task through sequence generation methodology.
    \item CoAD \cite{wang2023coad}, which integrates symptom and disease generation with label alignment for a comprehensive diagnostic model.
\end{itemize}

\subsection{Selection of Backbone Model}
Considering the language of automatic diagnosis datasets utilized, our selection for the backbone model is the Baichuan2-chat, a large language model with 13 billion parameters \cite{yang2023baichuan}. This model demonstrates robust performance in both languages. We also evaluate the multiple-choice symbol binding ability of the model using the metric of mean Proportion of Plurality Agreement (PPA) values \cite{robinson2023leveraging}, with $n$ answer options across the automatic diagnosis datasets. The results ranged from 0.78 to 0.94, indicating a notable consistency in the model performance regardless of the order of answer options, thereby affirming its reliability in MCQA tasks.

\subsection{Implementation of Diagnostic Framework}
For the agent-specialist of our AMSC framework, we employ the corresponding output logits of the answer option index of the last token in the last layer of Baichuan2 to formulate a diagnostic distribution for potential diseases. In our adaptive probability distribution fusion mechanism, the input dimension, denoted as $d_{input}$, is calculated as the product of the number of potential diseases and the number of agent specialists. The dimensions of the query, key, and value matrices ($q, k, v$) are set to $d_{input} \times d_{input}$. The learning rate for this implementation ranges between 1e-1 and 1e-3.

\subsection{Main Results}
We present an empirical evaluation of the AMSC framework, comparing its performance against established baselines across three real-world datasets. This evaluation encompasses several key metrics: diagnostic accuracy, training duration (denoted as Ttime), and the average number of turns required for implicit symptom inquiry (referred to as Turn). Additionally, we introduce a novel metric, the number of training parameters (\#Params), to further assess the efficiency of the automatic diagnosis system.

Table \ref{main_result} delineates the comparative performance of the AMSC framework and the baseline methodologies. In terms of diagnostic accuracy, the AMSC framework demonstrates superior efficacy, outperforming existing systems on all three datasets, with an enhancement in performance of up to 3.2\%. Regarding training time, since the LLM is frozen and only the APDF module requires training, the AMSC framework exhibits a significant reduction, saving up to 95.5\% training time compared with the baseline models, equating to a mere 0.09 minutes on the DXY dataset. Furthermore, the demand for AMSC for training parameters is notably efficient, scaling proportionally with the number of diseases. For a dataset encompassing $n$ diseases, the AMSC requires $3n^4+n^3$ training parameters, which signifies a reduction of up to 99.99\% in parameter training. Concerning the mean turn of symptom inquiry, the AMSC relies solely on explicit symptoms, obviating the need for implicit symptom inquiries. This approach is substantiated by experimental findings, affirming the feasibility of an automatic diagnosis framework operating without the integration of implicit symptoms.

By now, the AMSC framework has not incorporated implicit symptoms in its diagnostic process. However, prior work \cite{chen2022diaformer,chen2023dxformer,wang2023coad} suggests that implicit symptoms could enhance the performance of automatic diagnosis systems, which is also intuitive. Therefore, we will explore how the implicit symptoms will affect the performance of our AMSC framework in the following section. 

\section{Discussion}
\subsection{Impact of Implicit Symptoms}
\begin{table}[hb]
\centering
\renewcommand{\arraystretch}{1.1}
\begin{tabular}{lccc}
\hline
\textbf{Model} & \textbf{LLM} & \textbf{Single agent}  & \textbf{AMSC} \\ \hline
AMSC\dag                   & 0.632 & 0.676 & 0.773     \\  \hline
\textit{implicit symptoms}                          &&& \\
\ \ \ \  -all               & 0.631 & 0.669 & 0.752     \\  
\ \ \ \  -pos               & 0.638 & 0.647 & 0.760     \\  \hline
\textit{medical knowledge}                  &&&         \\
\ \ \ \  -reordered\dag    & N/A     & 0.641 &  0.718    \\  
\ \ \ \ \ \ \ -all          & N/A     & 0.662 &  0.724    \\  
\ \ \ \  -irrelevant\dag        & N/A     & 0.585 &  0.702    \\  
\ \ \ \ \ \ \ -all          & N/A     & 0.627 &  0.693    \\  \hline
\end{tabular}
\caption{Accuracy of LLM, Single agent, and AMSC with specific settings for the implicit symptoms and medical knowledge on the MuZhi-4 dataset. \dag: with no implicit symptom. -all: with both explicit and implicit symptoms. -pos: only with the positive symptoms in the explicit and implicit symptoms. N/A: not applicable.}
\label{w_imp_n_wrong_knowledge}
\end{table}
Experimental results have demonstrated that our AMSC framework can work well on the automatic diagnosis task even without the implicit symptoms. In this subsection, we will scrutinize how the implicit symptoms affect the performance of our AMSC with two settings on the MuZhi-4 dataset: 
\begin{itemize}
    \item all symptoms, denoted as ``-all'', includes entire explicit and implicit symptoms; 
    \item positive symptoms, denoted as ``-pos'', only involves the positive symptoms inside the explicit and implicit symptoms. 
\end{itemize}

Interestingly, our results, as outlined in Table \ref{w_imp_n_wrong_knowledge} reveal that the inclusion of a full spectrum of symptoms does not necessarily enhance the performance of AMSC framework. Also, configurations relying solely on positive symptoms yielded similar outcomes. The p-values of results of the paired t-test between the (AMSC\dag, AMSC -all) and (AMSC\dag, AMSC -pos) are 0.244 and 0.357 respectively, which means that there is \textit{no} statistically significant difference between the result with and without implicit symptom.

These findings have profound implications for the use of Large Language Models (LLMs) in medical diagnostics. In clinical practice, a comprehensive collection of symptoms is vital for accurate diagnosis. However, for LLMs, the abundance of symptoms, especially without differentiated weighting, poses interpretative challenges. Our analysis reveals that 59.1\% of symptoms are common across all four studied diseases in MuZhi-4, with decreasing proportions for three, two, and one disease of 25.8\%, 10.6\%, and 4.5\% respectively. This significant overlap complicates the task of accurate diagnosis with LLMs. For example, the symptom ``cough'' is present in 95.8\% of bronchitis cases and 53.1\% of Upper Respiratory Tract Infections (URTI). Furthermore, explicit symptoms, which are primarily patient-reported and considered dominant, can effectively reflect a condition of the patient to a specialist agent, demonstrating a notable aspect for LLM-based diagnostic systems.

\subsection{Impact of Medical Knowledge}
The medical knowledge has been proven to be helpful for the agent specialist with the experiment in Table \ref{main_result} and Figure \ref{recall_analysis}. Here we investigate the influence of medical knowledge on the performance of LLM-based automatic diagnosis systems from another angle. Utilizing the MuZhi-4 dataset, we explore two distinct strategies for medical knowledge:

\begin{itemize}
    \item Reordered Knowledge: We assign medical knowledge of one disease to another, creating mismatched knowledge for the LLMs. This is formulated as:
    \begin{equation*}
    k_n^{reordered} = d_n , (sym_{m_1},sym_{m_2}, \ldots)  
    \end{equation*}
    where $(sym_{m_1},sym_{m_2}, \ldots)$ are symptoms of disease $d_m$ different from the target disease $d_n$.
    
    \item Irrelevant Knowledge: Here, the automatic diagnosis system is provided with accurate medical knowledge of diseases not included in the automatic diagnosis task. It simulates a scenario where specialists unrelated to the specific diseases are consulted. This is represented as:
    \begin{equation*}
    k_o^{other} = d_o , (sym_{o_1},sym_{o_2}, \ldots)  
    \end{equation*}
    where $(sym_{o_1},sym_{o_2}, \ldots)$ are symptoms of a non-target disease $d_o$.
\end{itemize}

As shown in Table \ref{w_imp_n_wrong_knowledge}, the use of reordered or irrelevant medical knowledge adversely affects the diagnostic accuracy of both single-agent systems and AMSC configurations. While agents with reordered knowledge exhibit slightly better performance than those with entirely irrelevant knowledge, their effectiveness is significantly compromised due to the deliberate misalignment of disease-knowledge pairings.

\subsection{Decision Fusion Techniques}
We introduce and evaluate the Adaptive Probability Distribution Fusion (APDF) method, which is designed to amalgamate probability distributions of potential diseases, derived from a group of agent specialists. In parallel, we also conduct empirical evaluations with alternative fusion methodologies: (1) Majority Decision Fusion, wherein predominant decisions override minority opinions, (2) Mean Decision Fusion, which determines disease diagnoses based on the average distribution across multiple specialists, and (3) Linear Classifier Fusion. 

\begin{figure}[ht] 
\includegraphics[width=\columnwidth]{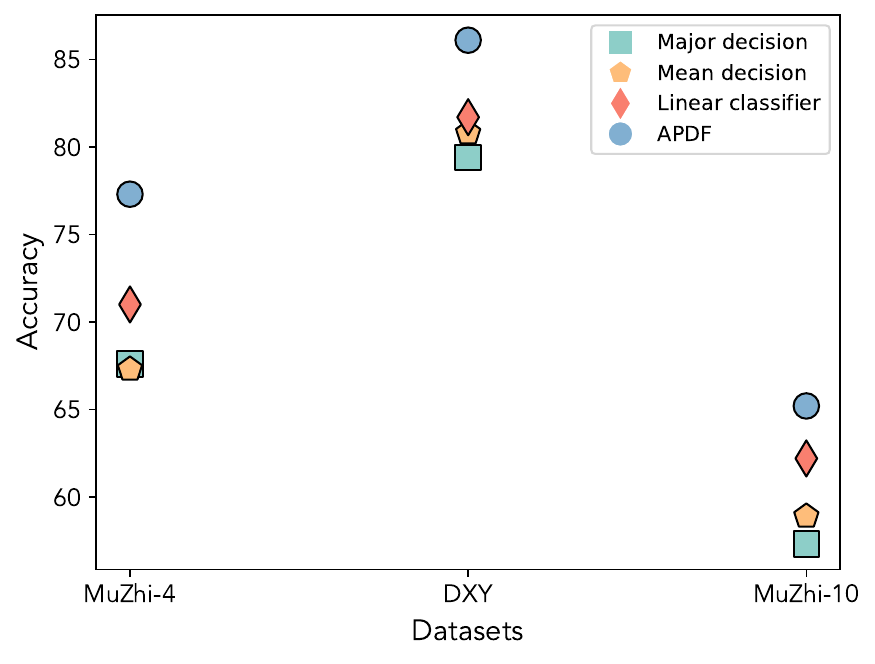} 
\caption{Accuracy for automatic diagnosis with various decision fusion techniques.} 
\label{fusion} 
\end{figure}

Figure \ref{fusion} represents the comparative accuracy achieved by these diverse decision fusion methods. It is observed that our proposed APDF method consistently surpasses the three baselines mentioned above approaches in terms of performance efficiency. Notably, the Linear Classifier Fusion demonstrates superior efficacy compared to both Majority and Mean Decision Fusions. The results above underscore the significant contribution of our APDF method to enhance the AMSC framework's effectiveness.

\subsection{Generalization on the Real-world Scenario}
Automated diagnosis datasets have converted patient-practitioner dialogues, which are inherently natural language-based, into normalized symptom labels to align with the methodologies established in prior work. However, a significant challenge arises in annotating the symptoms in the dialogues: it necessitates the involvement of highly skilled human experts, which renders the process cost-intensive and less feasible for real-world application. Our proposed framework AMSC, which utilizes a question template for symptom labels, can seamlessly integrate with natural language descriptions of symptoms. Notably, only the DXY dataset includes original self-reported symptom descriptions in natural language. Thus, we apply the AMSC to this dataset and compare the efficacy of the AMSC framework on normalized symptom labels and symptoms in natural language along with those of  ``Single agent'' and ``LLM''. The results are detailed in Table \ref{generalization}, which underscores the capability of AMSC to perform comparably with natural language symptom descriptions and demonstrates its potential for generalization in real-world scenarios.

\begin{table}[htbp]
\centering
\renewcommand{\arraystretch}{1.15}
\begin{tabular}{lccc}
\hline
& \textbf{LLM} & \textbf{Single agent} & \textbf{AMSC}   \\ \hline

Normalized          & 0.724 & 0.776 & 0.861       \\ 
Natural language    & 0.726 & 0.781 & 0.857       \\  \hline
\end{tabular}
\caption{Accuracy of automatic diagnosis with inputs of normalized symptom label and natural language for the LLM, Single agent and the AMSC on DXY dataset.}

\label{generalization}
\end{table}

\section{Conclusion}
In this paper, we employ agent-derived specialists for the automatic diagnosis tasks with open-source LLMs and medical knowledge for possible diseases. We introduce an agent-derived multi-specialist consultation framework, designed to assimilate and adaptively merge the probability distributions generated by these agent specialists in the diagnostic process, mimicking the real-life medical scene of a patient with multiple specialists. Experimental results demonstrate that our approach not only surpasses existing benchmarks in diagnostic accuracy but also significantly reduces the requisite training duration. Additionally, our in-depth analysis sheds light on an intriguing facet: the contribution of implicit symptom prediction toward diagnostic efficacy may be less substantial than previously anticipated. This revelation opens new avenues for exploration in the realm of automated diagnostic tasks.

\clearpage
\newpage

\section*{Ethics Statement}
This study is fundamentally an academic inquiry and is not designed for practical clinical diagnostic applications. The medical information employed herein is derived from publicly available medical knowledge repositories and reputable medical advisories. It must be underscored that diagnoses generated by LLMs should not be misconstrued as definitive. Should individuals encounter any form of discomfort or ailment, it is imperative to consult a certified healthcare practitioner.

\bibliographystyle{named}
\bibliography{ijcai23}

\end{document}